\pdfoutput=1

\documentclass[11pt]{article}

\usepackage[final]{acl}

\usepackage{times}
\usepackage{latexsym}

\usepackage[T1]{fontenc}

\usepackage[utf8]{inputenc}

\usepackage{microtype}

\usepackage{inconsolata}

\usepackage{amsmath,amsfonts,bm}

\def\eqref#1{equation~\ref{#1}}

\def\1{\bm{1}}

\DeclareMathAlphabet{\mathsfit}{\encodingdefault}{\sfdefault}{m}{sl}
\SetMathAlphabet{\mathsfit}{bold}{\encodingdefault}{\sfdefault}{bx}{n}

\usepackage{hyperref}
\usepackage{url}
\usepackage{graphicx}
\usepackage{subfigure}
\usepackage{subcaption}
\usepackage{tabularx}
\usepackage{multirow}
\usepackage{adjustbox}
\usepackage{booktabs}
\usepackage{mathtools}
\usepackage{longtable}
\usepackage{makecell}
\usepackage{hhline}
\usepackage{booktabs}
\usepackage{tablefootnote}
\usepackage{threeparttable}
\usepackage{wrapfig}
\usepackage[normalem]{ulem}
\usepackage{amssymb}
\usepackage{paralist}

\title{\textsc{PRoLoRA}: Partial Rotation Empowers More Parameter-Efficient LoRA }

\author{
\bf Sheng Wang$^{\heartsuit}$, 
Boyang Xue$^{\spadesuit}$, 
Jiacheng Ye$^{\heartsuit}$, 
Jiyue Jiang$^{\spadesuit}$,
Liheng Chen$^{\heartsuit}$, \\
\bf Lingpeng Kong$^{\heartsuit}$,
Chuan Wu$^{\heartsuit}$ \\
$^{\heartsuit}$ The University of Hong Kong, $^{\spadesuit}$ The Chinese University of Hong Kong \\
{\tt
u3009618@connect.hku.hk,
byxue@se.cuhk.edu.hk,
carsonye@connect.hku.hk, } \\
{\tt
jiangjy@link.cuhk.edu.hk, 
clh648@connect.hku.hk,
\{lpk, cwu\}@cs.hku.hk
}
}

\begin{document}

\maketitle

\begin{abstract}

With the rapid scaling of large language models (LLMs), serving numerous low-rank adaptations (LoRAs) concurrently 
has become increasingly impractical, leading to unaffordable 
costs and necessitating 
more parameter-efficient finetuning methods.
In this work, we introduce \textbf{P}artially \textbf{Ro}tation-enhanced \textbf{Lo}w-\textbf{R}ank \textbf{A}daptation (PRoLoRA), an intra-layer sharing mechanism comprising four essential components: broadcast reduction, rotation enhancement, partially-sharing refinement, and rectified initialization strategy.
As a superset of LoRA,
PRoLoRA retains its advantages,
and effectively circumvent the drawbacks of peer parameter-sharing methods with superior model capacity, practical feasibility, and broad applicability.
Empirical experiments demonstrate the remarkably higher parameter efficiency of PRoLoRA in both specific parameter budget and performance target scenarios, and its scalability to larger LLMs. Notably, with one time less trainable parameters, PRoLoRA still outperforms LoRA on multiple instruction tuning datasets.
Subsequently, an ablation study is conducted to validate the necessity of individual components 
and highlight the superiority of PRoLoRA over three potential 
variants.
Hopefully, the conspicuously higher parameter efficiency can establish PRoLoRA as a resource-friendly alternative to LoRA.

\end{abstract}

\section{Introduction} 

\begin{table}[!t]
\centering
\begin{tabularx}{\linewidth}{@{\extracolsep{\fill}} c @{\extracolsep{\fill}} c @{\extracolsep{\fill}} c @{\extracolsep{\fill}} c @{\extracolsep{\fill}} }
\toprule
Model                                               & \multirow{2}{*}{Rank} & \multicolumn{2}{c}{LoRA}                                   \\   
\cmidrule(){1-1} \cmidrule(l){3-4} 
LLaMA2                                              &                       & \multicolumn{1}{l}{Parameters} & \multicolumn{1}{l}{Bytes} \\ 
\midrule
\multirow{3}{*}{7B}                                 & 2                     & 5.00M                          & 19MB                     \\ 
                                                    & 16                    & 39.98M                         & 153MB                     \\ 
                                                    & 64                    & 159.91M                        & 610MB                       \\ 
                                                    \midrule
\multirow{3}{*}{13B}                                & 2                     & 6.26M                          & 24MB                       \\ 
                                                    & 16                    & 50.07M                         & 191MB                       \\ 
                                                    & 64                    & 200.28M                        & 764MB                       \\ 
                                                    \midrule
\multirow{3}{*}{70B}                                & 2                     & 11.27M                         & 43MB                       \\ 
                                                    & 16                    & 90.18M                         & 344MB                       \\ 
                                                    & 64                    & 360.71M                        & 1,376MB                      \\
\bottomrule
\end{tabularx}
\caption{Theoretical memory usage of LoRA weights with different ranks for LLaMA2-7/13/70B models.}
\label{tab: param count}
\end{table}




Finetuning large language models (LLMs), such as  LLaMA~2~\citep{Touvron2023a}, GPT-3.5~Turbo~\citep{OpenAI2023}, and Gemini~\citep{Team2023}, for specific domains and functions (e.g., model alignment~\citep{Wang2023b} and instruction tuning~\citep{Zhang2023a}
), have become increasingly popular.
To alleviate the high costs associated with full finetuning,
parameter-efficient finetuning (PEFT), especially LoRA~\citep{Hu2021}, has emerged as a lightweight solution by tuning a minority of parameters and freezing the remaining ones~\citep{Houlsby2019, Liu2022}.
However, with the rapid boost in the number of model's parameters, the demand for further enhanced parameter efficiency
becomes progressively more imperative, especially when multiple LoRA are deployed simultaneously.
As shown in the Table~\ref{tab: param count}, the configuration in~\citet{Dettmers2023} (i.e., applying LoRA  with the rank of 64 to all linear layers) results in a significant number of trainable parameters. For a single LLaMA2-7B model, LoRA will have about 160 million parameters to be tuned, occupying 610MB of disk storage and GPU memory in inference. These numbers quickly escalate to about 360 million and 1.4GB for a LLaMA2-70B model. For multi-LoRA scenarios, such as personalization~\citep{Chen2023a} and multitasking~\citep{Huang2023}, 
this issue will dramatically exacerbate. 
Specifically, resource consumption will increase linearly with personalized customization, which will further experience a quadratic growth when coupled with multitasking. 
Hence, the high costs in multi-LoRA scenarios do spark a demand for further improved parameter efficiency.

Focusing on the above target, parameter sharing can serve as an effective approach. Although small ranks could provide competitive performance in specific tasks ~\citep{Hu2021}, models generally perform better with higher ranks as listed in Table~\ref{tab: main results}. 
Besides, given a specific trainable parameter budget, better performance means higher parameter efficiency. Hence, enhancing parameter efficiency can be transformed into appropriately increasing the rank of LoRA with the same parameter count. 
Although VeRA~\citep{Kopiczko2023} can be regarded as an attempt in this direction, its aggressive freezing operations result in limited model capacity and excessively high rank, leading to significant inference latency in multi-LoRA scenarios where LoRA modules are not merged into the pretrained weights. 
Subsequently, Tied LoRA~\citep{Renduchintala2023} alleviates these problems by allowing the inter-layer shared matrices to be trainable. However, its tying mechanism restricts its applicability to weights of different shapes, which are widely present among self-attention and MLP modules.

To circumvent all the above drawbacks, we introduce a new approach called \textbf{P}artially \textbf{Ro}tation-enhanced \textbf{Lo}w-\textbf{R}ank \textbf{A}daptation (PRoLoRA). 
It features a parameter-sharing mechanism within the low-rank decomposition matrices, and consists of four essential components: broadcast reduction, rotation enhancement, partially-sharing refinement, and rectified initialization strategy. 
Specifically, we reparameterize the low-rank matrices with multiple chunks along the hidden dimension, and broadcast the first chunks to the others so that trainable parameters can be saved, or equivalently, the rank can be increased multiple times.
Then a nearly cost-free rotation operation along the rank dimension is performed to differentiate the identical chunks for higher expressiveness. 
Besides, a minimal subset of ranks is reserved without sharing for further refined capacity. To ensure the same bounds for initialization as unshared parameters, we also rectify the vanilla Kaiming uniform distribution~\citep{He2015} for shared ones.
As a superset of LoRA, PRoLoRA not only pertains the advantages of LoRA, such as lightweight task switching and optional merging to eliminate extra latency, but also brings about better capacity, practical feasibility, and broader applicability than other parameter-sharing methods.
Empirical experiments on multiple instruction tuning datasets validate the higher parameter efficiency of PRoLoRA than baselines via two alternative perspectives (i.e., a specific trainable parameter budget and performance target). With half of tunable parameters, PRoLoRA achieves 4/6 wins and better average performance over LoRA. When scaling up to LLaMA2-13B, PRoLoRA consistently outperforms LoRA with the same trainable parameter count.
Additionally, comprehensive ablation studies demonstrate the necessity of each component 
and the superiority of PRoLoRA over three potential intra-layer sharing variants.
Overall, PRoLoRA achieves significantly higher parameter efficiency, and thereby 
remarkably alleviates the storage and GPU memory burden in multi-LoRA scenarios, establishing PRoLoRA as a resource-friendly alternative for LoRA.

In summary, our main contributions are as follows: 
\begin{enumerate}[\textbullet]
    \item We introduce a more parameter-efficient method named PRoLoRA, featuring an intra-layer sharing mechanism consisting of broadcast reduction, rotation enhancement, partially-sharing refinement and rectified initialization strategy.

    \item We compare PRoLoRA with LoRA and existing peer methods on multiple instruction tuning datasets, and demonstrate its remarkably higher parameter efficiency, hopefully establishing PRoLoRA as a resource-friendly alternative to LoRA. 

    \item We perform an ablation study to demonstrate the necessity of individual components
    and the superiority of PRoLoRA over other potential
    variants.
\end{enumerate}

\section{Related work} 

\paragraph{LoRA Series.}
Inspired by the low intrinsic dimensions 
in over-parameterized models~\citep{
Aghajanyan2020},~\citet{Hu2021} proposes LoRA to reparameterize the weight update with two trainable low-rank matrices, while freezing the pretrained weights.
With this lightweight decomposition, LoRA reduces storage and task-switching overhead by sharing the pretrained models across multiple tasks.
Theoretically, the linear design of LoRA enables seamless merging of trainable matrices with frozen weights, thereby avoiding extra inference latency, albeit this operation is typically not performed in multiple LoRA serving scenarios.

Subsequently, numerous efforts have been made to further enhance the effectiveness and efficiency of LoRA. 
Based on singular value decomposition, AdaLoRA~\citep{Zhang2023} achieves the automatic rank allocation
by adaptively pruning less important parameters during finetuning, but the varying ranks among layers pose challenges on deploying multiple LoRAs.
Inspired by random projections~\citep{Aghajanyan2020}, VeRA~\citep{Kopiczko2023} shares two frozen random matrices across all layers, and updates disentangled combination vectors for each layer. Although this approach reduces the number of parameters, it results in performance degradation and several times the additional computation associated with a very high rank.
In contrast, Tied LoRA~\citep{Renduchintala2023} enhances parameter efficiency by sharing trainable LoRA matrices across all layers, with the down projection matrix further tied among the query, key, and value modules. Additionally, it incorporates scaling vectors to differentiate each module. 
However, in addition to the high rank similar to VeRA, this approach also necessitates the shared matrices to have identical shapes, which further restricts its expansion to other linear layers.
In contrast, our approach features an intra-layer sharing mechanism to enhance parameter efficiency, thereby circumventing the above drawbacks while exhibiting potential for integration with them.

\paragraph{Parameter Sharing.}
Parameter sharing has been adopted by prior studies to reduce model footprint.
Universal Transformer~\citep{Dehghani2018} proposes to share all layers within a transformer model. 
Systematically, \citet{Takase2023} refines this mechanism with three parameter-sharing strategies across transformer layers, improving the computational and parameter efficiency.
Similarly, \citet{Reid2021}~compares several parameter reduction methods and introduces the Subformer model with shared middle layers and embedding factorization, significantly saving the parameters without performance degradation.
Subsequently, DictFormer~\citep{Lou2021} reparameterizes the original weights with a shared dictionary, unshared coefficients and indices, resulting in a more compact transformer model and faster computations.
Targeting on-device deployment, 
EdgeFormer~\citep{Ge2022} shares the attention and FFN modules, and incorporates PEFT-based layer adaptation to minimize the number of parameters.
More recently, \citet{Pires2023} removes the FFN on the decoder layers and shares a single larger FFN across the encoder, achieving substantial gains in both accuracy and latency.
Differently, our study focuses on multi-LoRA scenarios and seeks to enhance the parameter efficiency of LoRA models instead of transformer models.

\begin{figure*}[!ht]
        \centering
        \hfill
        \subfigure[LoRA]{
            \includegraphics[width=0.205\linewidth]{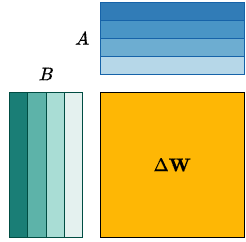}
            \label{fig: lora}
            }
        \hfill
        \subfigure[CLoRA]{
            \includegraphics[width=0.205\linewidth]{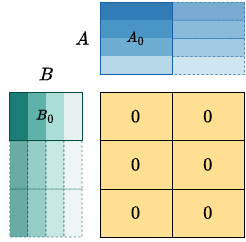}
            \label{fig: clora}
            }        
        \hfill
        \subfigure[RoLoRA]{
            \includegraphics[width=0.205\linewidth]{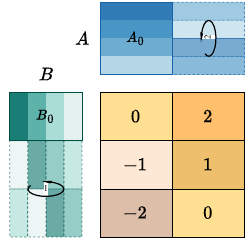}
            \label{fig: rolora}
            }
        \hfill
        \subfigure[PRoLoRA]{
            \includegraphics[width=0.205\linewidth]{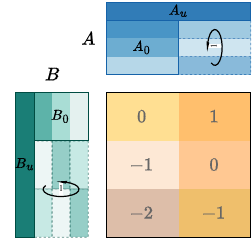}
            \label{fig: prolora}
            }
        \hfill
        \caption{
        Illustration of the original LoRA, our proposed PRoLoRA, and their intermediate states (i.e., CLoRA and RoLoRA). 
        Here we set the rank $r$, unshared rank $u$, sharing rates $m$ and $n$ of the $\mathbf{A}$ and $\mathbf{B}$ matrices to be 4, 1, 2 and 3, respectively. 
        Different shades of color in matrices $\mathbf{A}$ and $\mathbf{B}$ denote distinct ranks.
        The rotation arrows and center numbers indicate rotation directions and base strides, while dotted lines and higher transparency denote replicated or rotated weights, emphasizing that these weights do not contribute to the trainable parameters.
        Additionally, the center numbers 
        of each matrix block represent the relative displacement of the $\mathbf{A}_i$ and $\mathbf{B}_i$ chunks compared to those of top-left block (i.e., $\mathbf{A}_0$ and $\mathbf{B}_0$).
        }
        \label{fig: loras}
\end{figure*}

\section{Method} 
In this section, we introduce \textbf{P}artially \textbf{Ro}tation-enhanced \textbf{Lo}w-\textbf{R}ank \textbf{A}daptation (PRoLoRA), a more parameter-efficient method  featuring an intra-layer sharing mechanism.
Briefly, we present four essential components of PRoLoRA in Section~\ref{sec: prolora} namely broadcast reduction, rotation enhancement, partially-sharing refinement, and rectified initialization strategy, followed by its advantage analysis compared to existing peer methods in Section~\ref{sec: advantage}.

\subsection{
Mathematical Formulation
} \label{sec: prolora}
Based on the low-rank decomposition, LoRA~\citep{Hu2021} updates the frozen pretrained weights $\mathbf{W}_0 \in \mathbb{R}^{o \times h}$  with two trainable
matrices $\mathbf{A} \in \mathbb{R}^{r \times h}$ and $\mathbf{B} \in \mathbb{R}^{o \times r}$, where $r$ denotes the rank and satisfies $r \ll \min (h, o)$. This approximation process can be formulated as follows:
\begin{equation}
\mathbf{W}=\mathbf{W}_0+\Delta \mathbf{W}=\mathbf{W}_0+ \mathbf{BA}, 
\label{eq: lora}
\footnote{Here we omit the preset scaling factor $\frac{\alpha}{r}$ for clarity, but will restore it in Section~\ref{sec: exp}. }
\end{equation}
where $\mathbf{W}$ and $\Delta \mathbf{W}$ refer to the updated weights and weight differences with the dimensions of $o \times h$, respectively. 
Therefore, a higher parameter efficiency can be converted into how to obtain similar expressiveness of $\Delta \mathbf{W}$ with fewer parameters, which inspires the introduction of PRoLoRA.

\paragraph{Broadcast Reduction.}
An intuitive approach to optimizing the utilization efficiency of parameters is to reuse them multiple times. 
As depicted in Figure~\ref{fig: lora} and \ref{fig: clora}, in the first step of PRoLoRA, we propose to partition the original $\mathbf{A}$ and $\mathbf{B}$ matrices into chunks along the hidden dimensions $h$ and $o$ correspondingly, and broadcast the first chunks' parameters to the remaining ones, so that the expanded matrices maintain the same shapes as the original ones. 
Importantly, given the potential variations in the dimensions of $\mathbf{A}$ and $\mathbf{B}$, we allocate separate chunks to each matrix so that PRoLoRA will not be restricted by different weight shapes which is suffered by Tied LoRA.
Formally, this chunk-wise sharing process, referred to as CLoRA for simplicity, can be expressed as follows: 
\begin{equation}
    \begin{split}
        \Delta \mathbf{W} = 
        & (\underbrace{\mathbf{B}_0 \oplus_v \mathbf{B}_0 \oplus_v \ldots \oplus_v \mathbf{B}_0}_\text{n}) \\
        & (\underbrace{\mathbf{A}_0 \oplus_h \mathbf{A}_0 \oplus_h \ldots \oplus_h \mathbf{A}_0}_\text{m}),
    \end{split}
\label{eq: clora}
\end{equation}
where $\mathbf{A}_0$ and $\mathbf{B}_0$ refer to the trainable chunks shared $m$ and $n$ times in $\mathbf{A}$ and $\mathbf{B}$,  
while the symbols $\oplus_h$ and $\oplus_v$ denote the concatenation of two matrices horizontally and vertically, respectively. In this way, trainable parameters in one module can be reduced from the entire matrices $\mathbf{A}$ and $\mathbf{B}$ to two much smaller chunks $\mathbf{A}_0$ and $\mathbf{B_0}$, and the number decreases from $hr + ro$ to $hr/m + ro/n$. When $m$ equals to $n$, this broadcast reduction results in $m$ times fewer trainable parameters, or equivalently, $m$ times higher rank with the same trainable parameter budget, which potentially achieves better performance and parameter efficiency.

\paragraph{Rotation Enhancement.}
Despite the substantial decrease in tunable parameters, chunk-wise broadcast reduction also entails a trade-off in terms of the constrained expressiveness of the weight update. 
As illustrated in Figure~\ref{fig: clora}, when the matrices A and B are partitioned into the same chunks, their matrix product $\mathbf{BA}$ (i.e., $\Delta \mathbf{W}$) is also divided into multiple identical blocks along the row and column dimensions.
This notable pattern implies additional compression on the potential representational space of the weight differences, which intuitively might be detrimental to the performance.

To alleviate this issue, we further evolve CLoRA into RoLoRA, another intermediate transition approach towards PRoLoRA. As depicted in Figure~\ref{fig: rolora}, RoLoRA differentiates the broadcast chunks by rotating them with distinct times of a base stride.
Consequently, the computation of $\Delta \mathbf{W}$ can be transformed into the following form:
\begin{equation}
    \begin{split}
        \Delta \mathbf{W}= 
        & (\mathbf{B}_0 \oplus_v \mathbf{B}_1 \oplus_v \ldots \oplus_v \mathbf{B}_{n-1}) \\
        & (\mathbf{A}_0 \oplus_h \mathbf{A}_1 \oplus_h \ldots \oplus_h \mathbf{A}_{m-1}),
    \end{split}
\label{eq: rolora}
\end{equation}
where 
$\mathbf{A}_i$ and $\mathbf{B}_i$ are generated by applying the roll operation $\operatorname{Roll}()$ along the rank dimension to $\mathbf{A}_0$ and $\mathbf{B}_0$, respectively, with a base stride $s_{A}$ and $s_{B}$ multiplied by $i$. Mathematically, this can be expressed as $\mathbf{A}_i=\operatorname{Roll}(\mathbf{A}_0, i \cdot s_{A})$, and $\mathbf{B}_i=\operatorname{Roll}(\mathbf{B}_0, i \cdot s_{B})$, where $i$ ranges from $0$ to $m-1$ and $n-1$, correspondingly.
For simplicity and symmetry, we set $s_{A}$ and $s_{B}$ to
$\operatorname{Max}(\lfloor\frac{r}{m}\rfloor, 1)$ and $\operatorname{Max}(\lfloor\frac{r}{n}\rfloor, 1)$, respectively.
Since the rotation operation does not introduce any additional parameters, RoLoRA cost-freely enhances the expressiveness of CLoRA while preserving the same analysis of trainable parameters.

\paragraph{Partially-Sharing Refinement.}
Despite successfully avoiding simple replication, RoLoRA remains susceptible to a more subtle pattern.
Specifically, if two vectors simultaneously rotate in the same direction with a specific stride, their inner product keeps unchanged. The elements of $\Delta \mathbf{W}$ are computed from the inner product of corresponding column and row vectors in the matrices $\mathbf{A}$ and $\mathbf{B}$. Therefore, two blocks can still be identical if they are computed using chunk pairs with the same relative displacement.
For instance, as shown in Figure~\ref{fig: rolora}, the bottom right block is obtained by rotating both $\mathbf{A}_0$ and $\mathbf{B}_0$ with a stride of 2, resulting in an identical matrix product to the top left block. 
From an alternative perspective, while each block within a row/column is unique, blocks in different rows/columns can be derived by rotating the preceding row/column. Specially, if the base stride for rows and columns remains consistent (i.e., $s_{A}=s_{B}$), the resulting $\Delta \mathbf{W}$ exhibits a block-wise anti-diagonal symmetry. 
Despite being more implicit than that of CLoRA, this pattern still may hamper the performance of RoLoRA.

To refine the expressiveness of $\Delta \mathbf{W}$, we further introduce partially sharing on top of RoLoRA, leading to PRoLoRA. In detail, when partitioning the initial matrices $\mathbf{A}$ and $\mathbf{B}$ into chunks, a specific number of ranks, denoted as $u$, remain unshared.
By retaining independent hidden dimensions, these rank vectors are not restricted by the above implicit patterns, thereby 
allowing for refinements in the weight difference matrix $\Delta \mathbf{W}$ and an enriched representational capacity of PRoLoRA. 
To summarize, the whole scheme can be modeled as follows:
\begin{equation}
    \begin{split}
        \Delta \mathbf{W} = 
        & (\mathbf{B}_u \oplus_h (\mathbf{B}_0 \oplus_v \ldots \oplus_v \mathbf{B}_{n-1})) \\
        & (\mathbf{A}_u \oplus_v (\mathbf{A}_0 \oplus_h \ldots \oplus_h \mathbf{A}_{m-1})),
    \end{split}
\label{eq: prolora}
\end{equation}
where $\mathbf{A}_u \in \mathbb{R}^{u \times h}$ and $\mathbf{B}_u \in \mathbb{R}^{o \times u}$ are the unshared parts of $\mathbf{A}$ and $\mathbf{B}$, correspondingly. Different from the $\mathbf{A}_i$ and $\mathbf{B}_i$ in Eq.~\ref{eq: rolora}, the rank dimension of $\mathbf{A}_i$ and $\mathbf{B}_i$ here is $r-u$.
The introduction of partially sharing mechanism changes the trainable parameters, which now includes both the unshared and shared parts. The total number of trainable parameters in a module is given by $u(h+o) + h(r-u)/m + o(r-u)/n$. 
Due to the sharing mechanism, with a given trainable parameter count, PRoLoRA can still enjoy a higher rank, better performance and therefore higher parameter efficiency than LoRA.

\paragraph{Rectified Initialization Strategy.}
Following the default configuration of LoRA in the Huggingface PEFT v0.6.2 library~\citep{Mangrulkar2022}, we apply the vanilla Kaiming uniform initialization~\citep{He2015} to the unshared part $\mathbf{A}_u$.
However, due to the distinct fan-in dimensions of $\mathbf{A}_u$ and $\mathbf{A}_0$, Kaiming initialization inherently assigns them different sampling bounds, even if they collectively form the complete matrix $\mathbf{A}$.
Hence, we utilize the rectified Kaiming uniform initialization,
as formulated in Eq.~\ref{eq: kaiming}, for the shared chunk 
$\mathbf{A}_0$ to ensure unified bounds.
In contrast, the whole matrix $\mathbf{B}$ is initialized to be zero so that $\Delta \mathbf{W} = \mathbf{BA}$ is zero at the beginning of training, following the typical practice of LoRA~\citep{Hu2021}.
\begin{equation}
\mathbf{A}_0 \sim \mathcal{U}(-g \times \sqrt{\frac{3}{h}}, g \times \sqrt{\frac{3}{h}}),
\label{eq: kaiming}
\end{equation}
where $\mathcal{U}()$ denotes a uniform distribution, and $g$ means the gain determined by the non-linearity. Importantly, the hidden dimension $h$ of matrix $\mathbf{A}$ instead of that of chunk $\mathbf{A}_0$ is used to ensure the same bounds as the initialization of $\mathbf{A}_u$.

\subsection{Advantage Analysis} \label{sec: advantage}
As an intra-layer sharing mechanism, PRoLoRA would degrade to LoRA if the sharing mechanism is canceled (i.e., $u=r$). In other words, PRoLoRA can be regarded as a superset of LoRA with the same rank. Hence, PRoLoRA reserves various advantages of LoRA.
For example, PRoLoRA allows low-cost switching among tasks by swapping only the tunable weights
instead of all the parameters
on the fly, which is crucial for efficiently serving multiple customized models simultaneously. 
Notably, PRoLoRA keeps the linear property, and thereby can also be optionally merged into the pretrained weights in inference to eliminate extra inference latency.
Besides, PRoLoRA offers the following additional benefits compared to LoRA and other peer methods.

\paragraph{High Parameter Efficiency.}
As stated in Section~\ref{sec: main results} and~\ref{sec: llama13B}, PRoLoRA achieves better performance than LoRA and others given a specific parameter budget, indicating the apparently higher parameter efficiency of PRoLoRA. 
In other words, for a desired performance, PRoLoRA requires less disk space during storage,
and lower GPU memory during inference, which significantly alleviates the burden of serving multiple models.

\paragraph{High Representation Capacity.}
With the continuous increase of $r$, the performance of PRoLoRA can approximately converge to that of full fine-tuning as does LoRA. 
This guarantees a large model capacity, which is essential for challenging tasks. In comparison, the inferior performance of Tied LoRA and VeRA with the rank of 256 indicates their extremely limited capacity, as well as higher inference latency.


\paragraph{Broad Applicability.}
Different from Tied LoRA and VeRA, which share matrices across layers, PRoLoRA is an intra-layer sharing mechanism, thereby ensuring independence from the shape of pretrained weights. 
Specifically, when accounting for weights with varying shapes in the self-attention and MLP modules, VeRA requires initializing distinct shared matrix pairs for them, whereas the tying mechanism of Tied LoRA is no longer applicable.
In contrast, PRoLoRA enables decoupled weight sharing across all modules and is unaffected by different shapes, preserving the same level of applicability as LoRA. It even permits distinct unshared ranks and sharing ratios across layers.




\section{Experiments} \label{sec: exp}  
\subsection{General Setup} \label{sec: setup}
Overall, we focus on instruction-following tasks, and adhere to the settings of~\citet{Wang2023}. 
In particular, we similarly employ a multi-faceted assessment covering factual knowledge, reasoning, multilinguality, and coding, but carefully select the settings that yield positive effects based on the Table~7 in~\citet{Wang2023}.
We also convert all the datasets into a unified chatbot style, requiring models to learn both specific tasks and this interaction format. 
The core settings are presented as follows, while further details can be found in Appendix~\ref{sec: exp details}.

\paragraph{Datasets.}
    

To assess the \textbf{factual knowledge and multilinguality} abilities, we finetune models on Super-NaturalInstructions (SuperNI~\citep{Wang2022}) dataset and evaluate on Massive Multitask Language Understanding (MMLU~\citep{Hendrycks2021}) and TyDi~QA~\citep{Clark2020} datasets, respectively.
For the general and mathematical \textbf{reasoning}, we retrain the foundation models on Flan~V2 and its CoT split~\citep{Longpre2023}, and  report the performance on Big-Bench-Hard (BBH~\citep{Suzgun2022}) and the test split of Grade School Math (GSM~\citep{Cobbe2021}) corpora, respectively.
Moreover, we adopt HumanEval~\citep{Chen2021b} dataset to evaluate models' \textbf{coding} capability,
targeting models finetuned on CodeAlpaca~\citep{Chaudhary2023} dataset.


\paragraph{Baselines.}
We compare PRoLoRA with LoRA and other existing parameter-sharing baselines.
\begin{enumerate}[\textbullet]
    \item \textbf{LoRA}~\citep{Hu2021} adds trainable low-rank matrix pairs in parallel to the pretrained weights, as mentioned in Section~\ref{sec: prolora}. We 
    apply LoRA to all linear layers in transformer blocks, namely query, key, value, output, up, gate, and down projection weights. The scaling factor $\alpha$ and dropout rate are set to 16 and 0.1, respectively.

    \item \textbf{VeRA}~\citep{Kopiczko2023} shares and freezes two randomly initialized low-rank matrices, but updates the decoupled scaling vectors. We also apply it to all the linear layers, share frozen VeRA weights of the same types across layers, but initialize weights of different types separately. 

    \item \textbf{Tied LoRA}~\citep{Renduchintala2023} shares the trainable low-rank matrices among all the  query, key, and value projection layers, further ties their down projection matrices, and updates the separate scaling vectors for differentiation.

\end{enumerate}



\subsection{Main Results} \label{sec: main results}

\begin{table*}[!hbt]
\centering
\resizebox{\textwidth}{!}{%
\begin{tabular}{cccccccccc}
\toprule
\multirow{4}{*}{Method}         & \multirow{4}{*}{Rank} & \multirow{4}{*}{Param.}     & MMLU                 & BBH                    & GSM                     & \multicolumn{2}{c}{TyDi~QA}           & HumanEval & \multirow{4}{*}{Avg.} \\ 
                                &                       &                             & (factuality)         & (reasoning)            & (reasoning)             & \multicolumn{2}{c}{(multilinguality)} & (coding)   &         \\ \cmidrule(l){4-9} 
                                &                       &                             & EM                   & EM                     & EM                      & F1                 & EM               & P@1        &         \\ 
                                &                       &                             & (0-shot)             & (3-shot, Direct)       & (8-shot, CoT)           & \multicolumn{2}{c}{(1-shot, GP)}      & (0-shot)   &         \\ \midrule
Vanilla   (chat)                & -                     & -                           & 41.18                & 0.00                   & 3.03                    & 17.40              & 0.10             & 0.64       & 10.39    \\
Vanilla (no-chat)               & -                     & -                           & 41.53                & 33.43                  & 15.47                   & 49.18              & 35.35            & 13.57      & 31.42    \\ \midrule
\multirow{3}{*}{LoRA}           & 2                     & 5.00M                       & 44.77                & 36.22                  & 26.28                   & 48.67              & 35.70            & 18.24      & 34.98   \\ 
                                & 8                     & 19.99M                      & 46.55                & 36.92                  & 31.11                   & 50.50              & 36.89            & 19.37      & 36.89   \\
                                & 16                    & 40.98M                      & \textbf{46.70}       & 36.43                  & \textbf{31.34}          & 50.97              & 37.64            & 18.73      & 36.97   \\ \midrule
VeRA                            & 256                   & 1.42M                       & 42.51                & 35.10                  & 22.69                   & 48.39              & 36.38            & 18.90      & 34.00   \\ \midrule
\multirow{2}{*}{Tied LoRA}      & 256                   & 4.60M                       & 43.79                & 35.61                  & 26.66                   & 49.80              & 36.98            & 17.88      & 35.12   \\ 
                                & 280                   & 4.99M                       & 44.36                & 35.76                  & 25.47                   & 50.16              & 37.15            & 18.68      & 35.26   \\ \midrule
CLoRA                           & 16/32                 & 19.99M                      & 46.23                & \underline{36.88}      & 29.64                   & 49.72              & 36.81            & 19.14      & 36.40   \\ \midrule
RoLoRA                          & 16/32                 & 19.99M                      & 46.29                & \underline{36.44}      & 30.15                   & 50.79              & 37.85            & 19.41      & 36.82   \\ \midrule
\multirow{3}{*}{PRoLoRA}        & 8                     & 5.00M                       & 45.85                & 36.45                  & 27.14                   & 49.94              & 36.59            & 18.96      & 35.82   \\ 
                                & 4/8                   & 5.00M                       & \underline{45.85}    & \underline{36.45}      & 27.57                   & \underline{49.94}  & \underline{36.59}& \underline{19.75}   & 36.03   \\ 
                                & 16/32                 & 19.99M                      & 46.65                & \underline{\textbf{37.33}}  & 30.86              & \textbf{51.55}     & \textbf{37.90}   & \textbf{20.91}      & \textbf{37.53}     \\ 
\midrule
\multirow{1}{*}{$\text{PRoLoRA}^\text{-r}$}   &	16/32	            &	19.99M	                  &	46.27	             & \underline{36.56}    	          & 28.76	                & 50.68	             & 37.26	        & 20.56     	& 36.68 \\
\midrule
\multirow{1}{*}{$\text{PRoLoRA}^\text{-i}$}   & 16/32                 & 19.99M                      & 46.52                & \underline{37.25}                  & 30.45                   & 51.10              & 37.82            & 19.91        & 37.18     \\

\bottomrule                         
\end{tabular}
}
\caption{Results of LLaMA2-7B with different methods on diverse instruction following datasets.
“Param.” and “Avg.” are the abbreviations of “Parameter Count” and “Average”, while the symbols $^{-r}$ and $^{-i}$ denote the ablation of the rotation enhancement and rectified initialization strategy, respectively. “4/8” and “16/32” means raising the rank to either 4 or 8, and 16 or 32, respectively. Underlined represents the optional higher ranks, while bold indicates the best result for each benchmark.}

\label{tab: main results}
\end{table*}

When comparing the parameter efficiency of multiple methods, it is essential to answer two sequential questions. The first question is whether one method surpasses others in terms of parameter efficiency. Subsequently, the magnitude of efficiency enhancement needs to be measured. Both questions can be respectively analyzed from two alternative perspectives of parameter efficiency, as explained below.

\paragraph{Specific Parameter Budget.} 
The first view involves comparing the performance of different methods with a fixed trainable parameter count, where better performance signifies higher parameter efficiency. 
To accentuate the disparities among various methods and avoid the 
bias 
incurred by
parameter redundancy, we opt for a capacity-constrained scenario, wherein 
a limited parameter budget of about 5.00M is allowed.
As shown in Table~\ref{tab: main results}, LoRA with the rank of 2 exhibits an average performance of 34.98, outperforms the vanilla model, but consistently underperforms those with more trainable parameters, 
indicating a compact model capacity without apparent redundancy. Among these baselines, Tied LoRA achieves slightly better average performance of 35.12,
verifying its higher parameter efficiency as stated in~\citet{Renduchintala2023}, 
whereas VeRA does not~\footnote{Here we do not claim that VeRA has lower parameter efficiency than LoRA, considering that the inferior performance may be attributed to fewer trainable parameters. However, in our preliminary experiments on the MMLU dataset, increasing the rank to 4096 does not yield noticeable improvements.}.
However, due to 128 times higher ranks than LoRA, both of them result in much more computation and latency, diminishing their feasibility for latency-sensitive applications. 
In contrast, with the exactly same budget as LoRA, PRoLoRA exihibits remarkably better performance both individually and on average, when the rank, unshared rank and shared rates are set to 8, 1 and 7, respectively. Besides, if we optimize these hyperparameters for each task (i.e., optionally raising the rank to 4 or 8), its average performance can be further enhanced to 36.03, surpassing LoRA by over one percent.
This highlights that PRoLoRA achieves higher parameter efficiency than LoRA, while keeping better practical feasibility than other baselines.

\paragraph{Specified Performance Target.}
The other perspective is to achieve a desired performance with fewer tunable parameters, thereby quantifying parameter savings.
We increase the rank of LoRA eightfold to 16, establishing a good performance target across all the benchmarks without excessive parameter redundancy.
With the parameter count as 19.99M (equivalent to that of LoRA with the rank of 8), we raise the rank of PRoLoRA to 16 or 32 optionally.
The results in Table~\ref{tab: main results} show that PRoLoRA achieves 4/6 wins over the target individually and an average improvement from 36.97 to 37.53. 
This highlights that PRoLoRA accomplishes the performance target with half of trainable parameters, exhibiting its twice as high parameter efficiency.
To illustrate this advantage
explicitly, assuming 20GB of GPU memory available for customized parameters in inference, LoRA can serve for about 512 objects (e.g., users), while PRoLoRA doubles this number to 1024 without performance degradation. This is a remarkable benefit for service providers in multi-LoRA scenarios.

\subsection{Scalability Analysis} \label{sec: llama13B}
\begin{table}[!hbt]
\centering
\resizebox{\linewidth}{!}{%
\begin{tabular}{cccccc}
\toprule
\multirow{1}{*}{Method}  & \multirow{1}{*}{Rank} & MMLU            & BBH                       & GSM                      & \multirow{1}{*}{Avg.} \\ 
\midrule
Vanilla (chat)           & -                     & 50.05           & 0.00                      & 2.12                     & 17.39   \\
Vanilla (no-chat)        & -                     & 52.04           & 39.67                     & 27.82                    & 39.84   \\
\multirow{1}{*}{LoRA}    & 2                     & 52.78           & 42.50                     & 36.47                    & 43.92   \\
\multirow{1}{*}{PRoLoRA} 
                          & 4/8                  & \textbf{53.27}  & \textbf{43.12}            & \textbf{38.72}           & \textbf{45.04}   \\
\bottomrule                         
\end{tabular}
}
\caption{Results of LLaMA2-13B with different methods on three instruction following benchmarks.
}
\label{tab: 13B results}
\end{table}

We further verify the scalability of PRoLoRA on LLaMA2-13B models. Here we adopt a capacity-constrained setting with the limited tunable parameter budget as 6.26M, which is equivalent to that of LoRA with the rank of 2.
As shown in Table~\ref{tab: 13B results}, 
the average performance of LoRA is 43.92, whereas PRoLoRA achieves an improvement of 1. by boosting the rank to 4 or 8 
with the same budget,
reaching an average metric of 45.04. Impressively, PRoLoRA also consistently outperforms LoRA for individual tasks.  In summary, similar to LLaMA2-7B, PRoLoRA exhibits better performance with the same budget on LLaMA2-13B, validating its higher parameter efficiency on larger LLMs.

\subsection{Ablation Study} \label{sec: ablation}
We conduct an ablation study to evaluate the impact of each component of PRoLoRA, and explore its potential variants. All subsequent experiments adopt the LLaMA2-7B model with a fixed trainable parameter budget of 19.99M on BBH benchmark.
The hyperparameters remain consistent with those outlined in the preceding sections, except for those specifically under investigation.

\paragraph{Unshared Rank.}\label{sec: unshared rank}


\begin{figure}[!ht]
    \centering
    \includegraphics[width=\linewidth]{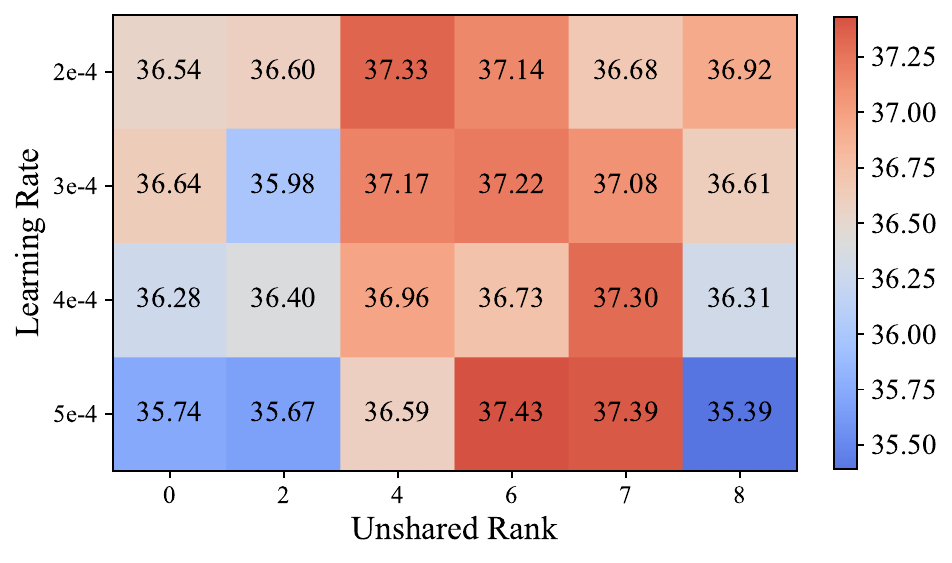}
    \caption{
    Performance of PRoLoRA with the rank of 32 with respect to unshared ranks and learning rates given a specific parameter budget on the LLaMA2-7B model and BBH benchmark. Specially, when the unshared rank is 8, all the ranks are unshared (i.e., vanilla LoRA).
    }
    \label{fig: unshared rank}
\end{figure}


We first study the impact of broadcast reduction and partially-sharing refinement on the performance. In Figure~\ref{fig: unshared rank}, we report the joint effects of unshared rank $u$ and learning rate. Specifically, with the unshared rank as 0, all trainable parameters are shared chunk-wisely, nullifying the partially-sharing mechanism. 
With the increase of the unshared rank, fewer parameters are shared, resulting in larger sharing ratios to keep the same rank (i.e., 32). Once the unshared rank reaches 8, all parameters are no longer shared, degrading PRoLoRA to vanilla LoRA.
Clearly, as the sharing ratios increase
with the presence of broadcast reduction,
the optimal learning rate gradually increases as well. 
This implies that shared and unshared parameters may require different learning rates, and setting them separately could further enhance PRoLoRA's performance, which is left for future work.
However, despite the unified learning rate, PRoLoRA consistently outperforms vanilla LoRA (i.e., $u=8$) with the unshared rank spanning from 4 to 7, demonstrating the superiority of increased rank through intra-layer sharing and the parameter robustness of PRoLoRA to the unshared rank. 
Besides, the inferior performance without unshared parameters highlights the necessity of partial sharing refinement in PRoLoRA.

\paragraph{Rotation Enhancement.}
We then investigate the enhancing effect of rotation on the representation capabilities of low-rank matrices. The results presented in Table~\ref{tab: main results} demonstrate the inferior performance of $\text{PRoLoRA}^{-r}$ (i.e., PRoLoRA without rotation enhancement ) across all the benchmarks, both individually and on average. This highlights the detrimental effect of explicit matrix patterns, resulting from simple duplication of chunks, on the expressiveness of PRoLoRA, as well as the nearly cost-free improvement brought by rotation. 

\paragraph{Initialization Strategy.}
We further examine the necessity of rectified initialization strategy.
As listed at the end of Table~\ref{tab: main results}, compared to PRoLoRA, $\text{PRoLoRA}^{-i}$, denoting PRoLoRA initialized with the vanilla Kaiming uniform distribution, consistently demonstrates inferior performance across all the tasks, resulting in an average performance drop to 37.18.
This suggests that initializing chunks directly, leading to larger sampling bounds, hinders the subsequent parameter optimization, and underscores the significance of the bound rectification.

\paragraph{Other Sharing and Rotations.} \label{sec: others}
\begin{table}[]
\centering
\begin{tabularx}{0.85\linewidth}{@{\extracolsep{\fill}} c @{\extracolsep{\fill}} c @{\extracolsep{\fill}} c @{\extracolsep{\fill}} }
\toprule
Sharing / Rotation & Hidden.    & Rank.     \\ \midrule
Hidden.           & 36.45 & \textbf{37.33} \\
Rank.             & 36.35 & 37.07 \\
\bottomrule
\end{tabularx}
\caption{Comparison of potential variants on LLaMA2-7B model and BBH benchmark. ``Hidden.'' and ``Rank.'' denote broadcast reduction or rotation enhancement along the hidden and rank dimensions, respectively.  }
\label{tab: variant}
\end{table}

Finally, we reaffirm the superiority of PRoLoRA by comparing it to three alternative intra-layer sharing mechanisms. Specifically, PRoLoRA is not the only possible approach that combines broadcast reduction and rotation enhancement. Both of these techniques can be applied along the hidden and rank dimensions, respectively. This yields four distinct combinations, in which PRoLoRA shares along the hidden dimension and rotates along the rank direction.
Table~\ref{tab: variant} presents the performance comparison of these approaches. Clearly, rotation enhancement along the rank direction achieves much better results than that along the hidden dimension, while broadcast reduction along the hidden dimension slightly outperforms that along the rank direction. PRoLoRA, incorporating these two favorable designs, achieves optimal performance, thereby establishing its superiority over other sibling variants.

\section{Conclusion}
Targeting more lightweight serving in multi-LoRA scenarios, we introduce PRoLoRA, a more efficient method featuring an intra-layer sharing mechanism consisting of broadcast reduction, rotation enhancement, partially-sharing refinement and rectified initialization strategy.
Empirically, we validate its higher parameter efficiency, scalability, and superiority over other methods, aiming to serve it as a resource-friendly alternative to LoRA.

\section{Limitation}

The limitations of this work mainly stem from the following two aspects:
\begin{enumerate}[\textbullet]
    \item As an intra-layer sharing mechanism, PRoLoRA may be potentially complemented by inter-layer sharing mechanisms, which is not covered by our current research. As shown in Table~\ref{tab: main results}, Tied LoRA, which leverages inter-layer sharing and is orthogonal to PRoLoRA, can slightly improve the parameter efficiency. This indicates that integrating both mechanisms may yield further improvements. Notably, PRoLoRA can adjust the hidden dimension of chunks as a common factor between the projection dimension of the self-attention mechanism and the intermediate dimension of the MLP module, before sharing chunks among them. This flexibility relieves the constraint of weight shape identity imposed by Tied LoRA.
    Nonetheless, exploring the feasibility of such integration necessitates a more comprehensive study that surpasses the scope of this paper, and is left for future work.
    
    \item As discussed in Section~\ref{sec: ablation}, setting separate learning rates for shared and unshared parameters may further improve the performance of PRoLoRA. However, we do not apply separate learning rates in our implementation, despite the potential additional benefits for PRoLoRA. As a remedy, grouping parameters with different learning rates in the optimizer or adding independent scalars, inspired by LoRA, to the unshared parameters hold promise for extra performance enhancement of PRoLoRA and warrant investigation in future studies.

    \item We claim that PRoLoRA requires lower GPU memory during inference in Section~\ref{sec: advantage}, even if our implementation of broadcasting allocates new memory to store the copied chunks, thereby negating any memory reduction for one module. However, we argue that it does not imply that overall memory usage cannot be reduced.
    Specifically, there are lots of PRoLoRA matrices in a model, and they are not computed simultaneously. When computing the preceding modules, the subsequent PRoLoRA matrices do not need to be copied. Once the computation of the preceding modules is completed, their occupied memory can be released promptly. Compared to the doubled parameter efficiency, we believe that the memory occupied by temporary copies can be negligible. In multi-user scenarios, only the active PRoLoRA matrices for active users need to be expanded, while other PRoLoRA matrices can still save lots of memory. 
    Besides, we also call for the efficient implementation of PRoLoRA with CUTLASS \footnote{https://github.com/NVIDIA/cutlass} and other relevant techniques.


\end{enumerate}








\section{Ethics Statement}
We strictly follow the ACL Code of Ethics during the research. To the best of our knowledge, there are no foreseeable potential risks in the methods we introduced. We report the computing infrastructure for all computational experiments presented in the paper. The transparent statistics on our results and detailed configuration of our experimental setup, including best-found hyperparameter values, are well stated. 
Besides, we will also release the code upon publication for publicly available reproducibility with minimal effort.
test~\citep{brown2020language}

\section{Acknowledgement}
This work was supported in part by Hong Kong Innovation and Technology Support Programme Platform Research Project fund (ITS/269/22FP), the joint research scheme of the National Natural Science Foundation of China (NSFC) and Hong Kong Research Grants Council (RGC) (under grant N\_HKU714/21), and RGC grants 17204423 and C7004-22G (CRF).

\bibliography{reference}

\newpage
\appendix

\section{Experiment Details} \label{sec: exp details} 
In our experiment implementation, we concentrate on instruction following tasks, and follow the settings outlined by \citet{Wang2023} as closely as possible. 
Similarly, we adopt a multi-faceted assessment, including factual knowledge, reasoning, multilinguality, and coding. Concurrently, we carefully select the settings that have demonstrated remarkably positive effects, as presented in Table~7 of \citet{Wang2023}. Further details are elucidated below.

\subsection{Dataset Details}
To assess different aspects of models' capabilities, we conduct specific evaluations with various datasets. 
Specifically, we finetune models on the Super-NaturalInstructions (SuperNI~\citep{Wang2022}) dataset, before reporting their performance on the Multitask Language Understanding (MMLU~\citep{Hendrycks2021}) for factual knowledge evaluation and on the TyDi~QA~\citep{Clark2020} dataset for multilingual ability.
To evaluate general and mathematical reasoning, we retrain the foundation models on the Flan~V2 dataset and its CoT split~\citep{Longpre2023}, respectively. We then present their corresponding performance on the Big-Bench-Hard (BBH~\citep{Suzgun2022}) and the test split of Grade School Math (GSM~\citep{Cobbe2021}) corpus.
Additionally, we employ the HumanEval~\citep{Chen2021b} benchmark to evaluate models' coding capability, targeting models finetuned on the CodeAlpaca~\citep{Chaudhary2023} dataset. 

In order to unify the diverse styles and formats, all the instruction tuning datasets are standardized to a chatbot-style schema. This involves the addition of two special tokens, namely <|user|> and <|assistant|>, preceding user utterances and assistant (i.e., language model) responses, respectively. Besides, another special token </s> is added to mark the end of each utterance or response. During training, only the sequences after the <|assistant|> token and before the subsequent <|user|> token are utilized for loss computation. Please refer to Figure~1 of \citet{Wang2023} for an illustrative example.

\paragraph{Finetuning Datasets.}
All the finetuning datasets used in our work are described as follows:
\begin{enumerate}[\textbullet]
    \item \textbf{SuperNI}~\citep{Wang2022} corpora encompasses a wide range of NLP tasks associated with instructions, and adheres to the Apache-2.0 license.

    \item \textbf{Flan~V2}~\citep{Longpre2023} dataset consolidates multiple pre-existing NLP datasets and enriches them with diverse data augmentations following~\citet{Chung2022}. The resulting mixture is made available under the Apache-2.0 license.

    \item \textbf{CoT}~\citep{Longpre2023} incorporates the annotation of chain-of-thoughts~\citep{Wei2022}. In accordance with \citet{Wang2023}, we utilize the CoT mixture extracted from the Flan~V2 dataset. 

    \item \textbf{CodeAlpaca}~\citep{Chaudhary2023} dataset is specifically designed for code generation, which is created with the Alpaca method and released under the Apache-2.0 license.
\end{enumerate}
\paragraph{Evaluation Datasets.}
Five multi-faceted evaluation benchmarks and their metrics deployed in our work are elucidated below.
\begin{enumerate}[\textbullet]
    \item \textbf{MMLU}~\citep{Hendrycks2021} is a benchmark designed to measure the factual knowledge capability of models. It comprises a collection of multiple-choice questions covering 57 subjects across STEM, humanities, social sciences, and more. The difficulty levels of these questions range from elementary to professional. In our evaluation, we employ the zero-shot setting and report the exact match (EM) score on it.

    \item \textbf{GSM}~\citep{Cobbe2021} corpora aims to assess the multi-step mathematical reasoning ability. It consists of 8.5K high-quality grade school math problems, including 1K for test, meticulously crafted by human writers. These problems involve a series of elementary arithmetic operations and require 2 to 8 steps to solve. We evaluate models on the test set of GSM with 8-shot examples and chain-of-thoughts (CoT), and report the EM score of the last number in the models' responses.

    \item \textbf{BBH}~\citep{Suzgun2022} is a suite of 23 challenging tasks taken from BIG-Bench~\citep{Srivastava2022}, designed to test the general multi-step reasoning abilities of language models. These tasks are specifically selected based on previous evaluations, where prior language models failed to surpass the average human-rater. Our evaluation utilizes 3 official few-shot examples without chain-of-thought (Direct), and reports the EM score accordingly. 

    \item \textbf{TyDi~QA}~\citep{Clark2020} is a multilingual question-answering dataset that includes 204K question-answer pairs in 11 topologically diverse languages and are collected directly in each language without any translation. It serves as a benchmark for evaluating models’ multilinguality performance. We adopt the gold passage (GP) setting where a gold passage containing the the correct answer is given as a reference, employ one-shot prompting and report both EM and F1 scores.

    \item \textbf{HumanEval}~\citep{Chen2021b} is a dataset containing 164 handwritten programming problems, each including a function signature, docstring, body, and several unit tests. It serves as a benchmark for evaluating models' coding capabilities by measuring the functional correctness for synthesizing programs from docstrings. Following the original paper, we report the pass@1 metric with zero-shot prompting and a sampling temperature of 0.1.
\end{enumerate}

\subsection{Hyperparameter Configurations}
We deploy LLaMA2-7B and 13B~\citep{Touvron2023a} as the foundation models, and conduct all the experiments on a single NVIDIA A100-40G GPU. Besides, all the settings are repeated three times with random seeds 1, 2, and 3, respectively, before the average performance is reported. Specific details for finetuning and evaluation are further explained below, respectively.

\paragraph{Finetuning Setup.}

To reduce the memory usage during finetuning, we follow QLoRA~\citep{Dettmers2023} to load all the pretrained models in 4-bit NormalFloat format and adopt a Paged AdamW Optimizer.
We then apply LoRA, VeRA, or PRoLoRA to all the linear layers in transformer blocks, including query, key, value, output, up, gate, and down projection weights, while setting the scaling factor $\alpha$ to 16 and the dropout rate to 0.1.
For more efficient finetuning, we group samples by length with a batch size of 16, and set the maximum sequence length to 512, which truncates samples if necessary.
We also disable weight decay and set the maximum gradient norm to 0.3 for better training stability.
Each model is finetuned for 10k steps with a linear learning rate scheduler and a warmup ratio of 3\%. 
Besides, we search for the optimal learning rate for each task. In detail, with the LoRA rank as 64, we search for the best learning rate among \{1e-5, 2e-5, 5e-5, 1e-4, 2e-4, 5e-4, and 1e-3\}, before the optimal values are fixed for both LoRA and PRoLoRA, unless otherwise stated~\footnote{Apparently, the learning rates optimized for LoRA may introduce slight unfairness to PRoLoRA. Even so, PRoLoRA still exhibits higher parameter efficiency, as presented in Section~\ref{sec: main results}.}. Our pre-experiments demonstrate that 5e-5 performs the best on the HumanEval benchmark, while 2e-4 outperforms the others on the remaining benchmarks.
To ensure a fair comparison, we further search for the optimal learning rates for VeRA and Tied LoRA with an extended range of \{2e-3, 5e-3, 1e-2, 2e-2, and 5e-2\}, to exhibit their best performance.

\paragraph{Evaluation Setup.}
During inference, we employ vLLM~\citep{Kwon2023}, which extremely accelerates the generation process with negligible impact on performance, and greedy decoding with a maximum length of 512. 
For the MMLU benchmark, as an exception, we load the models in an 8-bit format and set the evaluation batch size to 16.

\end{document}